\pdfoutput=1
\documentclass[11pt]{article}

\PassOptionsToPackage{a4paper,margin=2.5cm}{geometry}
\RequirePackage{geometry}

\usepackage{natbib}
\usepackage{times}
\usepackage{latexsym}
\usepackage[utf8]{inputenc}
\usepackage{microtype}
\usepackage{inconsolata}
\usepackage{amsmath}
\usepackage{graphicx}
\usepackage{caption}
\usepackage{hyperref}
\usepackage{xcolor}
\usepackage{ulem}
\usepackage{float}
\usepackage{graphicx}
\usepackage{tabularx}
\usepackage{array}
\usepackage{longtable}
\usepackage{multicol}
\usepackage{multirow}
\usepackage[english]{babel}
\usepackage{soul}
\usepackage{lipsum}
\usepackage{authblk}
\newenvironment{Figure}
  {\par\medskip\noindent\minipage{\linewidth}}
  {\endminipage\par\medskip}
\newenvironment{Table}
  {\par\medskip\noindent\minipage{\linewidth}}
  {\endminipage\par\medskip}
\title{Fool Me, Fool Me: User Attitudes Toward LLM Falsehoods}

\author{Diana Bar-Or Nirman, Ariel Weizman, Amos Azaria}
\affil{School of Computer Science, Ariel University, Israel}
\date{}

\begin{document}

\maketitle

\begin{multicols*}{2}
\begin{abstract}
While Large Language Models (LLMs) have become central tools in various fields, they often provide inaccurate or false information. This study examines user preferences regarding falsehood responses from LLMs. Specifically, we evaluate preferences for LLM responses where false statements are explicitly marked versus unmarked responses and preferences for confident falsehoods compared to LLM disclaimers acknowledging a lack of knowledge. Additionally, we investigate how requiring users to assess the truthfulness of statements influences these preferences.

Surprisingly, 61\% of users prefer unmarked falsehood responses over marked ones, and 69\% prefer confident falsehoods over LLMs admitting lack of knowledge. In all our experiments, a total of 300 users participated, contributing valuable data to our analysis and conclusions. When users are required to evaluate the truthfulness of statements, preferences for unmarked and falsehood responses decrease slightly but remain high. These findings suggest that user preferences, which influence LLM training via feedback mechanisms, may inadvertently encourage the generation of falsehoods. Future research should address the ethical and practical implications of aligning LLM behavior with such preferences.

\end{abstract}

\section{Introduction}
Large Language Models (LLMs) affect many aspects of our lives: programmers use them to obtain code snippets, students rely on them for homework assistance, and LLMs also play a significant role in literacy. People frequently use LLMs as a source of information in various fields, including exact sciences, life sciences, and history. The widespread use of LLMs as sources of information underscores the importance of ensuring that they generate true and accurate information. However, LLMs often generate inaccurate and even false information, which greatly impede their reliability as trusted tools for knowledge dissemination. This issue is particularly concerning given the confident tone and authoritative style in which LLMs present their outputs, making it difficult for users to differentiate between accurate information and inaccuracies. The persuasive nature of LLM-generated content increases the risk of misinformation, leading users to believe and propagate falsehoods.

Although it is widely assumed that users prefer accurate and truthful information, to the best of our knowledge, no prior work has explicitly tested this assumption in the context of LLMs. That is, no previous work has systematically investigated user preferences by presenting them with LLM-generated responses, where the veracity is disclosed a priori, and requiring users to evaluate or choose between truthful and false outputs. In this paper, we introduce a novel focus on user behavior in scenarios where the distinction between truth and falsehood is unambiguous. Specifically, we investigate user attitudes toward LLM falsehoods in two key issues:
\begin{enumerate}
    \item \textbf{Marked vs. Unmarked.} We examine user preferences between two types of LLM-generated responses: one where truth and falsehood are explicitly marked for easy distinction, and another with standard unmarked responses, where truth and falsehood appear identical in text format.
    \item \textbf{Uninformative Truth vs. Falsehood.} In scenarios where an LLM cannot provide the truth (e.g., when asked about events occurring after its cutoff date), we examine user preferences between two types of responses: one that acknowledges a lack of knowledge (uninformative truth) and another that provides a confident but inaccurate answer (falsehood).
\end{enumerate}

To explore user preferences, we conducted two experiments for each issue:
\begin{enumerate}
    \item A one-phase experiment, in which the two versions of ChatGPT-generated responses are represented to the participants and they should only select their preferred version.
    \item A two-phase experiment, in which after they select their preferred version in the first phase, the other version is hidden and participants must indicate whether a sentence in the ChatGPT response is true or false.
\end{enumerate}
The one-phase experiment examines initial user preferences, and the two-phase experiment examines the effect of taking responsibility on the user choice.

In the first issue, a clear majority of participants chose the unmarked version during the one-phase experiment. However, in the two-phase experiment, where users were required to take responsibility for their choice, preferences were nearly evenly split between the two versions. For the second issue, in both experiments, users overwhelmingly favored a confident but incorrect response over one acknowledging a lack of knowledge.

These findings are surprising as they challenge the assumption that individuals naturally favor truthful information over inaccurate responses. Previous research on user preferences regarding truth and falsehood has primarily examined behavior in contexts where the truth is ambiguous, such as studies on user trust in AI systems~\citep{bach2024systematic} and the dissemination of disinformation~\citep{buchanan2020people}. In contrast, this study explicitly presented the veracity of each statement. Despite this clarity, users still preferred the unmarked and falsehood versions.

Our findings have significant ethical and practical implications for LLM development. While users express a theoretical preference for transparency and accuracy, their real-time choices often gravitate toward aesthetically appealing but inaccurate responses.
Real-time choices may affect an LLM's performance, as user preferences are incorporated into its development through reinforcement learning from human feedback (RLHF), which fine-tunes LLMs based on user preferences~\citep{ziegler2019fine}. Therefore, our findings raise a question that must be addressed by LLM developers: Does RLHF inadvertently encourage LLMs to generate inaccurate or false information? Moreover, presenting the truth often requires complex expression and acknowledging uncertainty, which can drive users to prefer simpler yet incorrect responses. It highlights an ethical conflict for LLM developers: Should they prioritize factual accuracy or cater to user preferences for confident and appealing responses, even at the expense of truth?

\section{Related Work}
We now provide an overview of previous research on human tendencies to spread falsehoods and mention methods for detecting and reducing LLM-generated lies and hallucinations. Finally, we discuss methods for incorporating human feedback into machine learning models. 

\subsection{Spreading Disinformation} 

Spreading disinformation on social media is a common phenomenon with significant implications for public discourse and decision-making. Assuming people spread disinformation since they believe it to be true, Cialdini~\citep{cialdini2007influence} suggests three main reasons that persuade people to believe and disseminate disinformation:
\textbf{Consistency.} People may share information that aligns with their past behaviors or beliefs, as it helps reinforce their worldview and maintain cognitive coherence.
\textbf{Consensus.} People may share information that they perceive to be widely accepted or supported by others, as social proof plays a powerful role in validating beliefs.
\textbf{Authority.} People are more likely to share information originating from sources they deem credible or authoritative, even if the content itself lacks accuracy.

Tom Buchanan~\citep{buchanan2020people} confirmed the first reason and further identified demographic factors, finding that younger, male, and less-educated individuals were more likely to spread disinformation. Similarly, Vosoughi et al.~\citep{vosoughi2018spread} demonstrated that false information spreads faster, deeper, and more broadly on social media than truthful information, largely due to its novelty and emotional appeal, which captures users' attention and encourages sharing.

Additionally, Pennycook and Rand~\citep{pennycook2019fighting} noted that cognitive laziness and the failure to engage in analytical thinking contribute to the spread of disinformation. They argue that promoting digital literacy and critical thinking skills can mitigate this issue by enabling users to discern between credible and false content.

In contrast to studies on spreading disinformation, which assume that people believe the disinformation to be true, our research examines user preferences when they can clearly distinguish between truth and falsehood.


\subsection{LLM Hallucinations Reduction} 
One of the challenges in LLM research is detecting and reducing LLM lies and hallucinations.
Huang et al.~\citep{huang2023survey} provide a survey on LLM hallucinations, focusing on three main issues: hallucination causes, hallucination detection and benchmarks, and hallucination mitigation. 
For hallucination mitigation, some methods treat the LLM as a black-box, focusing their efforts on prompt engineering to achieve more trustworthy responses~\citep{peng2023check,madaan2024self}.
Other methods fine-tune the LLM for reducing hallucination based on human feedback~\citep{bakker2022fine}. However, our results suggest that user preferences may encourage lies rather than reduce them.
Another approach suggests using the intermediate states of an LLM to detect and reduce lies~\citep{AzariaM23}.

\subsection{Learning from Human Feedback}
Human provided information, such as data labeling and model performance evaluation, plays a major role in machine learning.
Kirk et al.~\citep{KirkBVRH23} survey the existing approaches for learning from human feedback. They start with before and after the advent of LLMs, continue with summarizing current methods
for incorporating human feedback learning into LLMs (e.g., reinforcement learning fine-tuning, supervised fine-tuning, and pretraining), and end with some future challenges.

One common method for learning from human feedback is reinforcement learning from human feedback (RLHF), which is commonly used for LLMs fine-tuning according to user preferences~\citep{ziegler2019fine}. RLHF is widely studied in various aspects~\citep{kaufmann2023survey}, including its affect on LLM generalization and diversity~\citep{kirk2023understanding} and construction, analysis, and improvements of algorithms for RLHF in LLMs~\citep{chaudhari2024rlhf,sessa2024bond}.


\section{Experimental Design}\label{sec.exp.design}
We conducted four experiments, labeled A, B, C, and D, to investigate user preferences. All the experiments use ChatGPT responses~\citep{chatgpt}. These experiments were administered through Amazon Mechanical Turk, a platform widely recognized as a reliable source for human subject experiments~\citep{paolacci2010running}. Participants had to meet the following criteria: task approval rate greater than 99, number of tasks approved greater than 1000, and participant location is USA. We now describe the experiments in detail, with the results being analyzed in Sec.~\ref{sec.results}. Each participant who completed the task received a payment of \$0.50, with an average completion time of approximately three minutes, equivalent to an hourly wage rate of approximately \$10. Participants provided consent for participating in each experiment. All experiments received IRB approval.

\subsection{Experiment A}
In Experiment A, participants were presented with two ChatGPT responses, with the following instructions:
“You will be shown two interactions of a user with ChatGPT.
One is the standard ChatGPT response, while the other ChatGPT response includes marking of 'false', 'true' and subjective or non-fact statements. The marked ChatGPT uses the following marking:
\begin{itemize}
    \item Bold and Underlined:  \textbf{\underline{This is a true fact.}};
    \item Crossed-out Grey text: \sout{\textcolor{gray}{This is a false fact.}};
    \item Regular Text: This sentence is subjective/not a fact.
\end{itemize}
You are required to select the response that you prefer better." 


Participants were required to select their preferred response. Each participant was presented with five pairs of such two responses, with one appearing on the right and other on the left. The location of responses and the order of pairs were randomly sampled to avoid positional bias. An example of a response pair is shown in Figure~\ref{fig:experiment1}.

\begin{Figure}
    \centering
    \includegraphics[width=\linewidth]{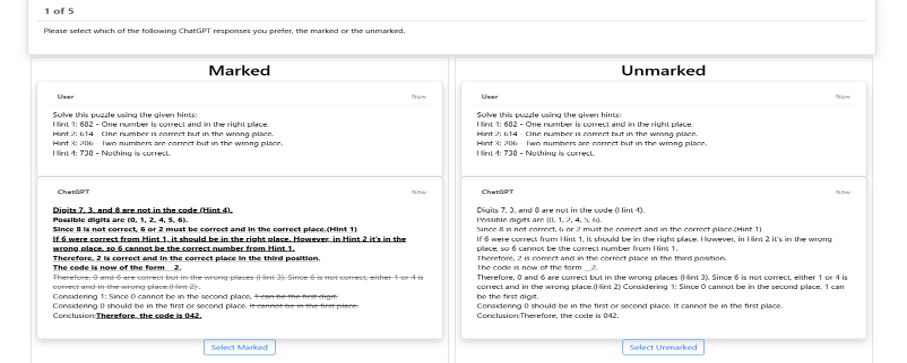}
    \captionof{figure}{Example of a response pair from Experiment A.}
    \label{fig:experiment1}
\end{Figure}

\subsection{Experiment B}
In Experiment B, we examined how requiring participants to verify the truth of sentences from a ChatGPT response after their initial choice influenced their preferences. This experiment consisted of two phases. In the first phase, participants were presented with a response pair as in Experiment A. After selecting their preferred response, the unselected version was removed, leaving only the chosen response visible.

In the second phase, participants were asked to determine whether a specific sentence from the ChatGPT's response is true, false or non-factual. If they provided an incorrect answer, they had to wait 30 seconds before attempting to correct it. Each participant completed such five two-phase surveys, ordered randomly. Figure~\ref{fig:experiment2} provides an example of the second phase, illustrating a marked response selection in the first phase.

When the unmarked version is selected, it may be more challenging for participants to answer the follow-up question correctly. This difficulty can activate the delay mechanism, requiring participants to wait 30 seconds before revising their answers. As a result, the delay mechanism may influence participants to choose the marked version in subsequent pair to avoid similar delays.

\begin{Figure}
    \centering
     \includegraphics[width=\linewidth]{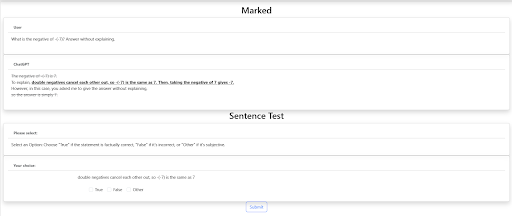}
    \captionof{figure}{Example of the second phase of Experiment B.}
    \label{fig:experiment2}
\end{Figure}




\subsection{Experiment C}
In Experiment C, participants were presented with two ChatGPT responses, with the following instructions: “You will be shown two interactions of a user with ChatGPT related to a recent event not known to ChatGPT.
In one interaction ChatGPT acknowledges that it may not know the answer, while in the other interaction it provides false information.
You are required to select the response that you prefer better." 


Each participant was presented with five pairs of such two responses, with one appearing on the right and other on the left. The location of responses (left and right) and the order of pairs were randomly sampled to avoid positional bias. An example of a response pair is shown in Figure~\ref{fig:experimentC}.

\begin{Figure}
    \centering
    \includegraphics[width=\linewidth]{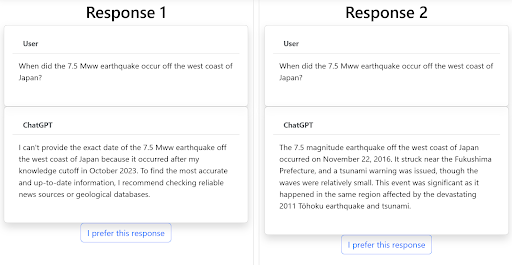}
    \captionof{figure}{Example of a response pair from Experiment C.}
    \label{fig:experimentC}
\end{Figure}

\subsection{Experiment D}
In Experiment D, we examined how requiring participants to verify the truth of sentences from a ChatGPT response after their initial choice influenced their preferences. This experiment consisted of two phases. In the first phase, participants were presented with a response pair as in Experiment C. After selecting their preferred response, the unselected version was removed, leaving only the chosen response visible.

In the second phase, participants were asked to determine whether a specific sentence from the ChatGPT's response is true or false. If they provided an incorrect answer, they had to wait 30 seconds before attempting to correct it. Each participant completed such five two-phase surveys, ordered randomly. Figure~\ref{fig:experiment4} provides an example of the second phase, illustrating an acknowledgment the lack of ChatGPT's knowledge selection.

When the unmarked version is selected, it may be more challenging for participants to answer the follow-up question correctly. This difficulty can activate the delay mechanism, requiring participants to wait 30 seconds before revising their answers. As a result, the delay mechanism may influence participants to choose the marked version in subsequent pair to avoid similar delays.

\begin{Figure}
    \centering
    \includegraphics[width=\linewidth]{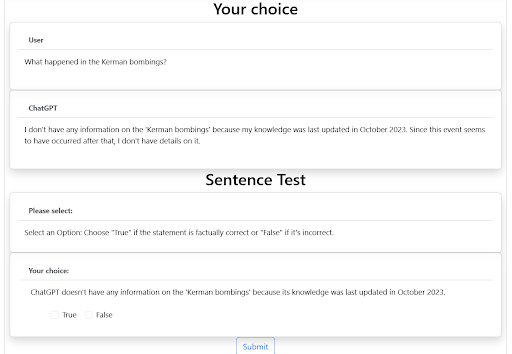}
    \captionof{figure}{An example of the second phase of Experiment D.}
    \label{fig:experiment4}
\end{Figure}

\section{Results}\label{sec.results}
We now present the results of the experiments. For all the experiments, we applied a $\chi^2$ test for goodness of fit. This test evaluates whether the observed differences in participant preferences between the two versions are statistically significant. The null hypothesis ($H_0$) assumes no difference in participant preferences between the two versions. Under this assumption, the expected values for each version are $\frac{N}{2}$, where $N$ represents the total number of surveys. The following sections detail the observed preferences and the corresponding statistical analyses for each experiment.

\subsection{Experiment A}
In Experiment A we recruited 74 participants, each of whom completed five surveys, totaling 370 surveys. About 60.1\% (223 out of 370) preferred the unmarked version, while only about 39.9\% (147 out of 370) preferred the marked version.

These results indicate a significant preference for the unmarked version. The observed values, 147 for marked and 223 for unmarked, yield a $\chi^2$ value of 15.6 with a $p$-value of $0.00008$. Since $p < 0.01$, we reject $H_0$, indicating a statistically significant preference for the unmarked version.

\subsection{Experiment B}
In Experiment B we recruited 75 participants, each of whom completed five surveys, totaling 375 surveys. About 51\% (192 out of 375) preferred the unmarked version, and about 49\% (183 out of 375) preferred the marked version.

Unlike Experiment A, in which there was a significant preference for the unmarked version, in Experiment B this preference is not significant. These results suggest a shift toward the marked version when the participants were required to verify the truth of specific sentences from their preferred version. The observed values, 183 for marked and 192 for unmarked, yield a $\chi^2$ value of 0.216 with a $p$-value of $0.6421$. Since $p > 0.05$ we fail to reject $H_0$, indicating no statistically significant difference in participants preferences.

\subsection{Experiment C}
In Experiment C we recruited 71 participants, each of whom completed five surveys, totaling 355 surveys. About 69.6\% (247 out of 355) preferred the falsehood response, while only about 30.4\% (108 out of 355) preferred the uninformative truth response.

These results indicate a significant preference for the falsehood response. The observed values, 247 for falsehood and 108 for uninformative truth, yield a $\chi^2$ value of 54.425 with a $p$-value less than 0.00001. Since $p < 0.01$, we reject $H_0$, indicating a statistically significant preference for the falsehood response.

\subsection{Experiment D}
In Experiment D we recruited 80 participants, each of whom completed five surveys, totaling 400 surveys. About 68.25\% (273 out of 400) preferred the falsehood response, while only about 31.75\% (127 out of 400) preferred uninformative truth.

These results indicate a significant preference for the falsehood response. The observed values, 273 for falsehood and 127 for uninformative truth, yield a $\chi^2$ value of 53.29 with a $p$-value less than 0.00001. Since $p < 0.01$, we reject $H_0$, indicating a statistically significant preference for the falsehood response.


\section{Additional Analysis}

\subsection{True/False Question for Participants}
In the second phase of Experiments B and D, participants were required to verify the truth of specific sentences from their preferred version. In Experiment B (Marked vs.~Unmarked), about 74\% (136 out of 183) of participants who preferred the marked version answered correctly, compared to about 70\% (134 out of 192) of those who preferred the unmarked version.

In Experiment D (Uninformative Truth vs. Falsehood), about 92\% (117 out of 127) of participants who preferred the uninformative truth response answered correctly, compared to about 85\% (232 out of 273) of those who preferred the falsehood response.

The success rates exceeding 70\% across all categories suggest that participants engaged seriously with the task, as random answers would have resulted in approximately 33\% success in Experiment B (as were three options: true, false, and subjective) and 50\% success in Experiment D (as were two options: true and false). Additionally, in both experiments, participants who selected the marked or uninformative truth responses answered correctly more often than those who selected the unmarked or falsehood responses. However, statistical analysis indicates that these differences are not significant, as the obtained $p$-values using the $\chi^2$ test were greater than $0.05$.

\subsection{Gender}
Table~\ref{table:gender} shows the difference between the choice of female and male. Using the $\chi^2$ test we assessed the statistical dependence between gender and the selected version. Statistical dependence between gender and choice was observed only in Experiment D. In Experiment B, males showed a higher preference for the marked version (54\%), while females preferred the unmarked version (55\%).
The statistical dependence observed in Experiment D may be attributed to differences in male and female behavior during the delay mechanism for incorrect answers in the second phase. In Experiment D, about 34.6\% of male participants who selected the falsehood response answered incorrectly, compared to only about 10.4\% of female participants. In contrast, in Experiment B, the gap between genders was smaller: about 31.6\% of male participants who selected the unmarked version answered incorrectly, compared to about 29.3\% of female participants.

\begin{Table}
    \centering
    \begin{tabular}{|c|c|c|c|c|c|}
		\hline
		& & & & &\\[-2.5ex]
		& Choice & F & M & $\chi^2$ & $p$-val.\\
		\hline
		\multirow{4}{*}{A}&\multirow{2}{*}{Marked}&&&\multirow{4}{*}{0.12}&\\[-5ex]\\
		& & 76 & 66 & &\\
        &&40\%&38\%&&0.73\\
        \cline{2-4}&&&&&\\[-5ex]\\
        &\multirow{2}{*}{Unmarked}&&&&\\[-5ex]\\
		& & 114 & 109 & &  $>0.05$\\
        &&60\%&62\%&&\\
		\hline
		\multirow{4}{*}{B}&\multirow{2}{*}{Marked}&&&\multirow{4}{*}{2.76}&\\[-5ex]\\
		& & 94 & 89 & &\\
        &&45\%&54\%&&0.097\\
		\cline{2-4}&&&&&\\[-5ex]\\
        &\multirow{2}{*}{Unmarked}&&&&\\[-5ex]\\
		&& 116 & 76 & & $>0.05$\\
        &&55\%&46\%&&\\
		\hline
        \multirow{4}{*}{C}&&&&\multirow{4}{*}{0.02}&\\[-5ex]\\
		& Uninf. & 72 & 175 & & \\
        & truth &69\%&70\%&& 0.89\\
		\cline{2-4}&&&&&\\[-5ex]\\
		& \multirow{2}{*}{Falsehood}&&&&\\[-5ex]\\
        && 33 & 75 & &$>0.05$ \\
        && 31\% &30\%&&\\
		\hline
        \multirow{4}{*}{D}&&&&\multirow{4}{*}{7.66}&\\[-5ex]\\
		& Uninf. & 84 & 43 &  & \\
        & truth &28\%&43\%&& 0.006\\
		\cline{2-4}&&&&&\\[-5ex]\\
        & \multirow{2}{*}{Falsehood}&&&&\\[-5ex]\\
		&& 221 & 57 &  &$<0.05$\\
        && 72\% & 57\%&&\\
		\hline
		\multicolumn{6}{c}{~}
	\end{tabular} \captionof{table}{Comparison between the choice of female (F) and male (M) in all of the experiments.}\label{table:gender}
\end{Table}


\subsection{Education Level}
We used the $\chi^2$ test to determine whether there is a statistical dependence between the education level of the users and their choices. Our findings indicate no statistical dependence between education level and user choices in Experiments A, B, and C. However, in Experiment D, a significant statistical dependence was observed, with a $\chi^2$ value of 81.4626 and a $p$-value less than 0.00001. Table~\ref{tab:eduLevel} provides a breakdown of user choices in Experiment D by education level.

\begin{Table}
    \centering
    \begin{tabular}{|c|c|c|c|}
        \hline
        &&&\\[-2.5ex]
        Choice & GS & Bachelor & HS \\
        \hline
        &&&\\[-2.5ex]
        Uninformative & 37 & 83 & 9 \\
        truth & 84\% & 24\% & 90\% \\
        \hline
        &&&\\[-2.5ex]
        \multirow{2}{*}{Falsehood} &&&\\[-5ex]\\
        & 7 & 267 & 1 \\
        & 16\% & 76\% & 10\%\\
        \hline
    \end{tabular}
    \captionof{table}{The choice of participants in Experiment D, partitioned according their education level.}
    \label{tab:eduLevel}
\end{Table}


\subsection{Feedback from the Participants}\label{sec:feedback}
To further explore participant preferences, a feedback mechanism was adjusted, including quantitative and verbal feedback. Table~\ref{tab:feedbackAB} presents the quantitative feedback for Experiments A and B, and Table~\ref{tab:feedbackCD} for Experiments C and D. The analysis of quantitative feedback reveals nuanced perspective on participant preferences when interacting with ChatGPT. In all experiments, participants rated the importance of accurate responses higher than that of confident but inaccurate responses.

For the verbal feedback, we asked the participants to provide comments about the versions they were presented with. A recurring theme in feedback for Experiments A and B was that the marked version is more convenient for fact-checking, whereas the unmarked version appears more aesthetically pleasing and neat. For Experiments c and D, participants commented that acknowledging a lack of knowledge increases their trust in ChatGPT.

These findings suggest that people, in principle, favor transparency and accuracy, indicating a strong preference for truthfulness as a guiding principle. However, when faced with real-time choices, people often gravitate toward more aesthetically appealing but potentially less accurate responses. This behavior illustrates the well-documented intention-behavior gap~\citep{sheeran2002intention}, similar to the dilemma of choosing between healthy and junk food~\citep{monds2016can,faries2016we}. While participants value transparency and accuracy in principle, their real-time choices often prioritize aesthetic appeal or convenience, much like individuals who express a preference for healthy food but choose junk food in practice.

\begin{Table}
\centering
\begin{tabular}{|c|>{\centering\arraybackslash}p{2.8cm}|>{\centering\arraybackslash}p{2.8cm}|} 
\hline
& &\\[-2.5ex]
& I believe that it is important for ChatGPT to mark which sentences are true and which are false. & I prefer that ChatGPT's responses will not include any marking.\\
\hline
& &\\[-2.5ex]
A & 3.81 & 3.74 \\ 
\hline
& &\\[-2.5ex]
B & 3.56 & 3.35  \\
\hline
\end{tabular}
\captionof{table}{Quantitative feedback from Experiments A and B (average, out of 5).}
\label{tab:feedbackAB}
\end{Table}

\begin{Table}
\centering
\begin{tabular}{|c|>{\centering\arraybackslash}p{2.8cm}|>{\centering\arraybackslash}p{2.8cm}|} 
\hline
& &\\[-2.5ex]
& I believe that it is important for ChatGPT to provide only accurate information. & I believe that it is important for ChatGPT to provide information even if it is incorrect. \\ 
\hline
& &\\[-2.5ex]
C & 4.11 & 3.17  \\ 
\hline
& &\\[-2.5ex]
D & 4.8 & 1.25  \\
\hline
\end{tabular}
\captionof{table}{Quantitative feedback from Experiments C and D (average, out of 5).}
\label{tab:feedbackCD}
\end{Table}

\section{Discussion}
In this paper we examined user preferences regarding LLM-generated falsehoods. Following Horton~\citep{horton2023large}, who suggests using LLMs as simulated economic agents, we explored this approach by conducting our four experiments with ChatGPT agents~\citep{chatgpt4o}. For each experiment, we simulated ChatGPT agents by replicating the demographic profiles (gender, age, and education level) of the actual participants and asked them to select their preferred version. As shown in Table~\ref{tab:comparisonGPTHuman}, ChatGPT agents exhibited a stronger inclination toward accurate responses (i.e., marked or uninformative truth version) compared to human participants. One conclusion from this is that human participants cannot be replaced by ChatGPT agents~\citep{dillion2023can}.

ChatGPT obtains human feedback by prompting users to select their preferred version between two options, as illustrated in Figure~\ref{fig:RLHF}. However, since our findings indicate that user preferences often incline toward inaccurate responses, directly incorporating these preferences into fine-tuning processes may not be advisable. Given that ChatGPT tends to favor accurate responses, it could incorporate a verification mechanism to assess the validity of user choices before using them for model fine-tuning. This verification could be performed by asking ChatGPT itself whether the preferred version is accurate. By leveraging its existing capabilities to evaluate factuality, ChatGPT could ensure that only reliable preferences are used for training.

Another approach to verification is for ChatGPT to conduct a preliminary test for each of the users. In this test, ChatGPT asks the user to select her preferred response between two responses, in which one response is an uninformative truth and the other is a falsehood. The user must be told a priory which response is truthful. Users who pass the test by selecting the truth will be marked by ChatGPT as reliable users, which their selections in any conversation can be used to fine-tune the model.
 However, this approach raises ethical concerns, as conducting such a verification test without the respondent's knowledge could violate principles of informed consent.

Alternatively, the regular feedback mechanism could be adapted to address this issue. ChatGPT could track the number of times each user selects a truthful option and use this data to create a reliability score. Feedback from users with high reliability scores could then be prioritized for fine-tuning, ensuring that training incorporates preferences aligned with factual accuracy while maintaining transparency and ethical standards.

\begin{Table}
\centering
\begin{tabular}{|c|>{\centering\arraybackslash}p{2.8cm}|>{\centering\arraybackslash}p{2.8cm}|} 
\hline
& &\\[-2.5ex]
& ChatGPT agents preferred the marked or uninformative truth version  & Human participants preferred the marked or uninformative truth version\\
\hline
& &\\[-2.5ex]
A & 45.3\% & 39.9\% \\ 
\hline
& &\\[-2.5ex]
B & 69\% & 49\%  \\
\hline
& &\\[-2.5ex]
C & 70.8\% & 30.4\%  \\
\hline
& &\\[-2.5ex]
D & 87.4\% & 31.75\%  \\
\hline
\end{tabular}
\captionof{table}{Comparison between ChatGPT agents and human participants preferences.}
\label{tab:comparisonGPTHuman}
\end{Table}

 \begin{Figure}
     \centering
     \includegraphics[width=\linewidth]{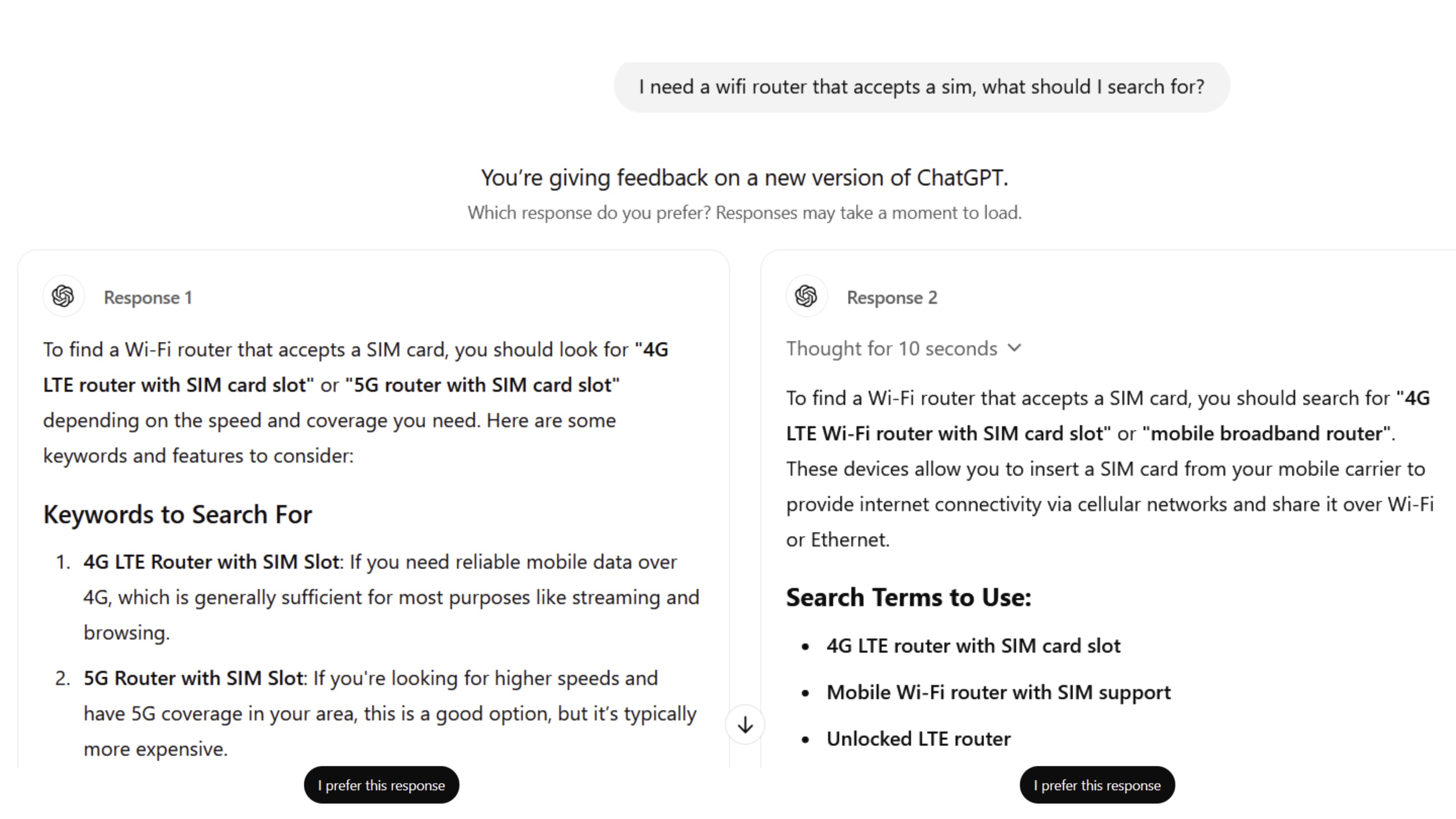}
     \captionof{figure}{Example of ChatGPT's human feedback mechanism.}
     \label{fig:RLHF}
 \end{Figure}

\section{Conclusion \& Future Work}
In this paper we discussed user preferences regarding LLMs-generated falsehoods, focusing on two key issues: (a) selecting between marked and unmarked versions, where the marked version allows distinguishing between truth and falsehood statements but it less aesthetically pleasing, and (b) selecting between uninformative truth and falsehood versions, where the falsehood is written with confidence, while the truth version acknowledges a lack of knowledge.

Our findings indicate that users tend to favor more aesthetically pleasing and confident responses, even at the expense of accuracy. This contradicts their fundamental agreement on truthfulness, as reflected in their feedback (see Section~\ref{sec:feedback}). Future work could explore alternative marking method, such as customizable or less accentuated designs, to investigate whether these approaches influence user preferences.

While users showed a strong preference for unmarked and falsehood versions, these preferences may vary for specific uses. For example, students and researchers using LLMs for their studies might value tools that help distinguish between truth and falsehood. Such tools could also be critical in healthcare applications, like fact-checking medical information for patients. Future work could examine user preferences within specific populations and use cases to better tailor LLM features.


\section{Limitations}
This paper examines user preferences regarding LLM-generated falsehoods through four experiments conducted on USA participants recruited via Amazon Mechanical Turk. However, our participant may not fully represent the broader population. Specifically, user preferences may vary across different cultural and linguistic contexts. Cultural and linguistic differences might influence how users perceive confidence, truthfulness, and markings. Examining the preferences of those who actively provide feedback to ChatGPT would have been valuable, but it is not feasible for us.

The experiments used pre-prepared ChatGPT responses, which limits the ability to capture user preferences in real-time interactions. Real-time interactions are advantageous because users prioritize topics they raise themselves, which may influence their preferences.

Conducting such experiments with user-generated interactions presents several challenges. One challenge is the need for an automatic marking tool that operates as efficiently as the original LLM. Additionally, the toll's error rate must be very low, as incorrectly marking truthful statements as falsehoods (or vice versa) could significantly influence user preferences. Furthermore, users may inquire about personal facts and perceive them as truthful, which contrasts with the rationale behind the marking system, as subjective statements should not be marked as truthful.

Finally, the binary choice framework used in this study (i.e., marked vs. unmarked, uninformative truth vs. falsehood) may oversimplify user preferences. Future research could explore more nuanced options, such as hybrid or customizable marking systems, to better understand user preferences.

\section{Ethical Statement}
This study was conducted in accordance with ethical research principles, ensuring respect for participant privacy and informed consent. Data was collected anonymously to protect participants' identities, and no personally identifiable information was stored or shared.
Additionally, the findings presented here aim to stimulate discussions on the ethical implications of aligning LLMs with human preferences, particularly when these preferences may increase LLM-generation of lies.
The authors are committed to advancing responsible AI development, particularly ensuring transparency and accuracy in LLMs. Furthermore, this paper raises critical ethical concerns that must be addressed, emphasizing the importance of balancing user preferences with the need for truthfulness and reliability in AI outputs.

\newpage
\bibliographystyle{acl_natbib}
\bibliography{references}

\begin{thebibliography}{24}
\expandafter\ifx\csname natexlab\endcsname\relax\def\natexlab#1{#1}\fi

\bibitem[{Azaria and Mitchell(2023)}]{AzariaM23}
Amos Azaria and Tom~M. Mitchell. 2023.
\newblock \href {https://doi.org/10.18653/V1/2023.FINDINGS-EMNLP.68} {{The Internal State of an {LLM} Knows When It's Lying}}.
\newblock In \emph{Findings of the Association for Computational Linguistics: {EMNLP} 2023, Singapore, December 6-10, 2023}, pages 967--976. Association for Computational Linguistics.

\bibitem[{Bach et~al.(2024)Bach, Khan, Hallock, Beltr{\~a}o, and Sousa}]{bach2024systematic}
Tita~Alissa Bach, Amna Khan, Harry Hallock, Gabriela Beltr{\~a}o, and Sonia Sousa. 2024.
\newblock {A systematic literature review of user trust in AI-enabled systems: An HCI perspective}.
\newblock \emph{International Journal of Human--Computer Interaction}, 40(5):1251--1266.

\bibitem[{Bakker et~al.(2022)Bakker, Chadwick, Sheahan, Tessler, Campbell-Gillingham, Balaguer, McAleese, Glaese, Aslanides, Botvinick et~al.}]{bakker2022fine}
Michiel Bakker, Martin Chadwick, Hannah Sheahan, Michael Tessler, Lucy Campbell-Gillingham, Jan Balaguer, Nat McAleese, Amelia Glaese, John Aslanides, Matt Botvinick, et~al. 2022.
\newblock {Fine-tuning language models to find agreement among humans with diverse preferences}.
\newblock \emph{Advances in Neural Information Processing Systems}, 35:38176--38189.

\bibitem[{Buchanan(2020)}]{buchanan2020people}
Tom Buchanan. 2020.
\newblock {Why do people spread false information online? The effects of message and viewer characteristics on self-reported likelihood of sharing social media disinformation}.
\newblock \emph{Plos one}, 15(10).

\bibitem[{Chaudhari et~al.(2024)Chaudhari, Aggarwal, Murahari, Rajpurohit, Kalyan, Narasimhan, Deshpande, and da~Silva}]{chaudhari2024rlhf}
Shreyas Chaudhari, Pranjal Aggarwal, Vishvak Murahari, Tanmay Rajpurohit, Ashwin Kalyan, Karthik Narasimhan, Ameet Deshpande, and Bruno~Castro da~Silva. 2024.
\newblock {RLHF Deciphered: A Critical Analysis of Reinforcement Learning from Human Feedback for LLMs}.
\newblock \emph{arXiv preprint arXiv:2404.08555}.

\bibitem[{Cialdini(2007)}]{cialdini2007influence}
Robert~B Cialdini. 2007.
\newblock \emph{{Influence: The psychology of persuasion}}, volume~55.
\newblock Collins New York.

\bibitem[{Dillion et~al.(2023)Dillion, Tandon, Gu, and Gray}]{dillion2023can}
Danica Dillion, Niket Tandon, Yuling Gu, and Kurt Gray. 2023.
\newblock {Can AI language models replace human participants?}
\newblock \emph{Trends in Cognitive Sciences}, 27(7):597--600.

\bibitem[{Faries(2016)}]{faries2016we}
Mark~D Faries. 2016.
\newblock {Why we don’t “just do it” understanding the intention-behavior gap in lifestyle medicine}.
\newblock \emph{American journal of lifestyle medicine}, 10(5):322--329.

\bibitem[{{Horton, John J}(2023)}]{horton2023large}
{Horton, John J}. 2023.
\newblock {Large language models as simulated economic agents: What can we learn from homo silicus?}
\newblock Technical report, National Bureau of Economic Research.

\bibitem[{Huang et~al.(2023)Huang, Yu, Ma, Zhong, Feng, Wang, Chen, Peng, Feng, Qin et~al.}]{huang2023survey}
Lei Huang, Weijiang Yu, Weitao Ma, Weihong Zhong, Zhangyin Feng, Haotian Wang, Qianglong Chen, Weihua Peng, Xiaocheng Feng, Bing Qin, et~al. 2023.
\newblock {A survey on hallucination in large language models: Principles, taxonomy, challenges, and open questions}.
\newblock \emph{ACM Transactions on Information Systems}.

\bibitem[{Kaufmann et~al.(2023)Kaufmann, Weng, Bengs, and H{\"u}llermeier}]{kaufmann2023survey}
Timo Kaufmann, Paul Weng, Viktor Bengs, and Eyke H{\"u}llermeier. 2023.
\newblock {A survey of reinforcement learning from human feedback}.
\newblock \emph{arXiv preprint arXiv:2312.14925}.

\bibitem[{Kirk et~al.(2023)Kirk, Bean, Vidgen, R{\"{o}}ttger, and Hale}]{KirkBVRH23}
Hannah Kirk, Andrew~M. Bean, Bertie Vidgen, Paul R{\"{o}}ttger, and Scott Hale. 2023.
\newblock \href {https://doi.org/10.18653/V1/2023.EMNLP-MAIN.148} {{The Past, Present and Better Future of Feedback Learning in Large Language Models for Subjective Human Preferences and Values}}.
\newblock In \emph{Proceedings of the 2023 Conference on Empirical Methods in Natural Language Processing, {EMNLP} 2023, Singapore, December 6-10, 2023}, pages 2409--2430. Association for Computational Linguistics.

\bibitem[{Kirk et~al.(2024)Kirk, Mediratta, Nalmpantis, Luketina, Hambro, Grefenstette, and Raileanu}]{kirk2023understanding}
Robert Kirk, Ishita Mediratta, Christoforos Nalmpantis, Jelena Luketina, Eric Hambro, Edward Grefenstette, and Roberta Raileanu. 2024.
\newblock \href {https://openreview.net/forum?id=PXD3FAVHJT} {{Understanding the Effects of {RLHF} on {LLM} Generalisation and Diversity}}.
\newblock In \emph{The Twelfth International Conference on Learning Representations, {ICLR} 2024, Vienna, Austria, May 7-11, 2024}. OpenReview.net.

\bibitem[{Madaan et~al.(2023)Madaan, Tandon, Gupta, Hallinan, Gao, Wiegreffe, Alon, Dziri, Prabhumoye, Yang, Gupta, Majumder, Hermann, Welleck, Yazdanbakhsh, and Clark}]{madaan2024self}
Aman Madaan, Niket Tandon, Prakhar Gupta, Skyler Hallinan, Luyu Gao, Sarah Wiegreffe, Uri Alon, Nouha Dziri, Shrimai Prabhumoye, Yiming Yang, Shashank Gupta, Bodhisattwa~Prasad Majumder, Katherine Hermann, Sean Welleck, Amir Yazdanbakhsh, and Peter Clark. 2023.
\newblock \href {http://papers.nips.cc/paper\_files/paper/2023/hash/91edff07232fb1b55a505a9e9f6c0ff3-Abstract-Conference.html} {{Self-Refine: Iterative Refinement with Self-Feedback}}.
\newblock In \emph{Advances in Neural Information Processing Systems 36: Annual Conference on Neural Information Processing Systems 2023, NeurIPS 2023, New Orleans, LA, USA, December 10 - 16, 2023}.

\bibitem[{Monds et~al.(2016)Monds, MacCann, Mullan, Wong, Todd, and Roberts}]{monds2016can}
Lauren~A Monds, Carolyn MacCann, Barbara~A Mullan, Cara Wong, Jemma Todd, and Richard~D Roberts. 2016.
\newblock {Can personality close the intention-behavior gap for healthy eating? An examination with the HEXACO personality traits}.
\newblock \emph{Psychology, health \& medicine}, 21(7):845--855.

\bibitem[{OpenAI(2024{\natexlab{a}})}]{chatgpt}
OpenAI. 2024{\natexlab{a}}.
\newblock \href {https://openai.com} {{ChatGPT 3.5 (June 16 version) [Large language model]. https://openai.com}}.

\bibitem[{OpenAI(2024{\natexlab{b}})}]{chatgpt4o}
OpenAI. 2024{\natexlab{b}}.
\newblock \href {https://openai.com} {{ChatGPT 4 (November 28 version) [Large language model]. https://openai.com}}.

\bibitem[{Paolacci et~al.(2010)Paolacci, Chandler, and Ipeirotis}]{paolacci2010running}
Gabriele Paolacci, Jesse Chandler, and Panagiotis~G Ipeirotis. 2010.
\newblock Running experiments on amazon mechanical turk.
\newblock \emph{Judgment and Decision making}, 5(5):411--419.

\bibitem[{Peng et~al.(2023)Peng, Galley, He, Cheng, Xie, Hu, Huang, Liden, Yu, Chen et~al.}]{peng2023check}
Baolin Peng, Michel Galley, Pengcheng He, Hao Cheng, Yujia Xie, Yu~Hu, Qiuyuan Huang, Lars Liden, Zhou Yu, Weizhu Chen, et~al. 2023.
\newblock {Check your facts and try again: Improving large language models with external knowledge and automated feedback}.
\newblock \emph{arXiv preprint arXiv:2302.12813}.

\bibitem[{Pennycook and Rand(2019)}]{pennycook2019fighting}
Gordon Pennycook and David~G Rand. 2019.
\newblock Fighting misinformation on social media using crowdsourced judgments of news source quality.
\newblock \emph{Proceedings of the National Academy of Sciences}, 116(7):2521--2526.

\bibitem[{Sessa et~al.(2024)Sessa, Dadashi, Hussenot, Ferret, Vieillard, Ram{\'{e}}, Shahriari, Perrin, Friesen, Cideron, Girgin, Stanczyk, Michi, Sinopalnikov, Ramos, H{\'{e}}liou, Severyn, Hoffman, Momchev, and Bachem}]{sessa2024bond}
Pier~Giuseppe Sessa, Robert Dadashi, L{\'{e}}onard Hussenot, Johan Ferret, Nino Vieillard, Alexandre Ram{\'{e}}, Bobak Shahriari, Sarah Perrin, Abe Friesen, Geoffrey Cideron, Sertan Girgin, Piotr Stanczyk, Andrea Michi, Danila Sinopalnikov, Sabela Ramos, Am{\'{e}}lie H{\'{e}}liou, Aliaksei Severyn, Matt Hoffman, Nikola Momchev, and Olivier Bachem. 2024.
\newblock {{BOND:} Aligning LLMs with Best-of-N Distillation}.
\newblock \emph{arXiv preprint arXiv:2407.14622}.

\bibitem[{Sheeran(2002)}]{sheeran2002intention}
Paschal Sheeran. 2002.
\newblock {Intention—behavior relations: a conceptual and empirical review}.
\newblock \emph{European review of social psychology}, 12(1):1--36.

\bibitem[{Vosoughi et~al.(2018)Vosoughi, Roy, and Aral}]{vosoughi2018spread}
Soroush Vosoughi, Deb Roy, and Sinan Aral. 2018.
\newblock The spread of true and false news online.
\newblock \emph{science}, 359(6380):1146--1151.

\bibitem[{Ziegler et~al.(2019)Ziegler, Stiennon, Wu, Brown, Radford, Amodei, Christiano, and Irving}]{ziegler2019fine}
Daniel~M Ziegler, Nisan Stiennon, Jeffrey Wu, Tom~B Brown, Alec Radford, Dario Amodei, Paul Christiano, and Geoffrey Irving. 2019.
\newblock {Fine-Tuning Language Models from Human Preferences}.
\newblock \emph{arXiv preprint arXiv:1909.08593}.

\end{thebibliography}
\end{multicols*}
\end{document}